\documentclass[10pt,twocolumn,letterpaper]{article}

\usepackage{cvpr}              

\definecolor{cvprblue}{rgb}{0.21,0.49,0.74}
\usepackage[pagebackref,breaklinks,colorlinks,allcolors=cvprblue]{hyperref}

\usepackage{multirow}
\usepackage{booktabs}
\usepackage{makecell}
\usepackage{pifont}

\def\paperID{5947} 
\def\confName{CVPR}
\def\confYear{2025}

\title{Enhanced Contrastive Learning with Multi-view Longitudinal Data for Chest X-ray Report Generation}

\author{\normalsize\textbf{Kang Liu}$^{1,2,3}$ \quad \normalsize\textbf{Zhuoqi Ma}$^{1,2,3,4*}$ \quad \normalsize\textbf{Xiaolu Kang}$^{1,2,3}$ \quad \normalsize\textbf{Yunan Li}$^{1,2,3}$\\
\normalsize\textbf{Kun Xie}$^{1,2}$ \quad \normalsize\textbf{Zhicheng Jiao}$^4$ \quad \normalsize\textbf{Qiguang Miao}$^{1,2,3}$\thanks{Corresponding author. The code is available at \url{https://github.com/mk-runner/MLRG}} \\
\normalsize$^{1}$School of Computer Science and Technology, Xidian University, Xi'an, China \\ 
\normalsize$^{2}$Xi'an Key Laboratory of Big Data and Intelligent Vision, Xi'an, China \\ 
\normalsize$^{3}$Key Laboratory of Collaborative Intelligence Systems, Ministry of Education, Xidian University, Xi'an, China \\ 
\normalsize$^{4}$Warren Alpert Medical School, Brown University, Providence, USA \\
\small\{kangliu, 22031212472\}@stu.xidian.edu.cn, \small\{zhuoqima, yunanli, xiekun, qgmiao\}@xidian.edu.cn, \small zhicheng\_jiao@brown.edu
}

\begin{document}
\maketitle
\begin{abstract}
Automated radiology report generation offers an effective solution to alleviate radiologists' workload. However, most existing methods focus primarily on single or fixed-view images to model current disease conditions, which limits diagnostic accuracy and overlooks disease progression. Although some approaches utilize longitudinal data to track disease progression, they still rely on single images to analyze current visits. To address these issues, we propose enhanced contrastive learning with \textbf{M}ulti-view \textbf{L}ongitudinal data to facilitate chest X-ray \textbf{R}eport \textbf{G}eneration, named MLRG. Specifically, we introduce a multi-view longitudinal contrastive learning method that integrates spatial information from current multi-view images and temporal information from longitudinal data. This method also utilizes the inherent spatiotemporal information of radiology reports to supervise the pre-training of visual and textual representations. Subsequently, we present a tokenized absence encoding technique to flexibly handle missing patient-specific prior knowledge, allowing the model to produce more accurate radiology reports based on available prior knowledge. Extensive experiments on MIMIC-CXR, MIMIC-ABN, and Two-view CXR datasets demonstrate that our MLRG outperforms recent state-of-the-art methods, achieving a 2.3\% BLEU-4 improvement on MIMIC-CXR, a 5.5\% F1 score improvement on MIMIC-ABN, and a 2.7\% F1 RadGraph improvement on Two-view CXR.
\end{abstract}    
\section{Introduction}
\label{sec:intro}

Chest X-ray (CXR) is a widely employed diagnostic tool in clinical practice, primarily for evaluating the lungs, heart, pleura, and skeletal structures. It is critical for diagnosing conditions, such as pneumonia, fracture, pneumothorax, pleural effusion, and cardiomegaly \cite{irvin-chexpert}. To ensure effective communication across departments and between physicians and patients, radiologists manually document detailed reports based on their interpretation of CXR images. However, this process is both expertise-dependent and time-consuming \cite{wang2024-survey-rrg,sei}. As the demand for imaging studies continues to grow, the workload associated with manual report generation may intensify, potentially impacting medical efficiency and compromising patient care quality \cite{2024-eccv-hergen,bannur2024maira2groundedradiologyreport}. To mitigate these challenges, radiology report generation (RRG) \cite{RBME-survey-rrg-2024,guo2024automaticmedicalreportgeneration-survey} has emerged as a promising solution. By automatically analyzing imaging data from X-ray \cite{liang2024divide-acmmm-24}, CT \cite{hamamci2024ct2rep}, or pathology \cite{guo2024histgen}, RRG generates clinical findings using factual terminology \cite{fse,yan2023style} and descriptive language. This automation aids radiologists by providing high-quality draft reports \cite{sei}, improving diagnostic efficiency.

\begin{figure}
    \centering
    \includegraphics[width=1\linewidth]{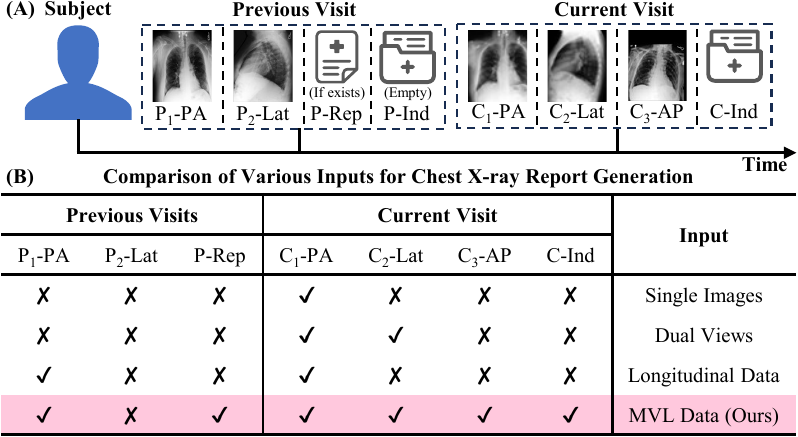}
    \caption{(A) shows medical historical data of a subject (patient) over time. (B) compares inputs for RRG, with \textit{AP} and \textit{PA} as frontal views, and \textit{Lat} and \textit{Rep} as a lateral view and its report. \textit{Ind} and \textit{MVL Data} are ``INDICATION" and multi-view longitudinal data.}
    \label{fig:1}
\end{figure}

In clinical practice, radiologists typically conduct comprehensive evaluations using multi-view images from the current visit, incorporate patient medical histories (i.e., longitudinal data) to track disease progression, and integrate patient-specific prior knowledge to assist in diagnosis and report generation. However, most existing RRG methods \cite{chen-etal-2020-generating,nicolson-improving-cvt2distilgpt2,tmm_mulview_2024-fmvp} focus solely on single images when generating reports and struggle to effectively differentiate between views, such as posteroanterior (PA), anteroposterior (AP), lateral, or left anterior oblique. These views exhibit inherent differences; for example, although both PA and AP views are frontal images, geometric variations can cause cardiac enlargement in the AP view, potentially impacting diagnostic accuracy. To address this issue, some studies \cite{chen-etal-2021-cross-modal,yang-m2kt} have introduced dual-view report generation, as illustrated in Figure \ref{fig:1}. Empirical results \cite{chen-etal-2020-generating,chen-etal-2021-cross-modal} reveal that incorporating dual views enhances the quality of generated reports. Nevertheless, these methods merely distinguish between frontal and lateral views, neglecting more subtle differences across multiple views. Moreover, both single-image and dual-view methods focus solely on the images from the current visit, disregarding the descriptions of disease progression found in radiology reports. This limitation may lead to model hallucinations. To combat this problem, some studies \cite{chexrelnet-longitudinal-miccai-2022-not-rrg,pre-fill-longitudinal-miccai-2023,2024-eccv-hergen} have sought to leverage longitudinal data to model disease progression, as shown in Figure \ref{fig:1}. However, these methods still rely on a single image to characterize the current visit, limiting diagnostic accuracy. Additionally, some patients may lack ``INDICATION", ``previous report", or ``previous image" due to their first visit or improper data storage. This variability challenges the flexible use of available data to generate accurate reports.

To mimic radiologists' diagnostic pipeline and address these challenges, we propose a two-stage MLRG for chest X-ray report generation. In Stage 1, the key part is our proposed multi-view longitudinal contrastive learning approach, which utilizes the inherent spatiotemporal information in radiology reports to supervise the pre-training of visual and textual representations. Specifically, we incorporate learnable position embeddings for each view to identify differences across varying numbers of views. We then employ a multi-view longitudinal fusion network that flexibly integrates spatial information from current multi-view images and temporal information from longitudinal data. Subsequently, we learn visual and textual representations by leveraging agreements between multi-view longitudinal data (see Figure \ref{fig:1}) and their corresponding radiology reports. In Stage 2, we introduce a tokenized absence encoding technique to handle missing patient-specific prior knowledge (i.e., ``INDICATION" and ``previous report"). This allows the multi-modal fusion network to adapt flexibly to the presence or absence of such data, improving the accuracy of the generated reports. Extensive experiments on MIMIC-CXR \cite{johnson-mimic-cxr-jpg}, MIMIC-ABN \cite{mimic-abn-ori}, and Two-view CXR \cite{mcl} datasets demonstrate the superiority of MLRG in producing clinically accurate reports. Our contributions are stated as follows:
\begin{itemize}
    \item We propose a novel multi-view longitudinal contrastive learning method that flexibly integrates multi-view longitudinal data and leverages the inherent spatiotemporal information from reports to supervise the pre-training of visual and textual representations.
    \item We introduce a tokenized absence encoding technique to handle missing patient-specific prior knowledge. This technique enables the model to adapt flexibly to scenarios with or without such data, ensuring the text generator can utilize available prior knowledge effectively.
    \item Our MLRG shows competitive results compared to various state-of-the-art methods across three public datasets: MIMIC-CXR, MIMIC-ABN, and Two-view CXR.
\end{itemize}


\section{Related Work}
\label{sec:related_work}

\textbf{Radiology report generation (RRG).} RRG is akin to image captioning \cite{li2023-blip2,cvpr-2023conzic-image-captioning}, but requires generating detailed content with specialized medical terminology. Existing RRG methods consist of a vision encoder (like ResNet101 \cite{chen-etal-2021-cross-modal,yang-m2kt,huang-kiut}, CvT \cite{nicolson-improving-cvt2distilgpt2,nicolson2023-longitudinal-multiview}, or ViT \cite{wang2023metransformer,liu2024in-context-acmmm}) and a text generator (such as Memory-driven Transformer \cite{chen-etal-2020-generating,fse}, MiniGPT-4 \cite{aaai-liu2024bootstrapping-llm}, DistilGPT2 \cite{nicolson-improving-cvt2distilgpt2,liang2024divide-acmmm-24}, or LLaMA \cite{liu2024in-context-acmmm,2024-iclr-cxr-llm,wang-2023-r2gengpt}). To improve clinical accuracy in RRG, researchers have incorporated various techniques or prior knowledge, including knowledge graphs \cite{zhang-when-aaai-2020,yang-gsket}, cross-modal alignment \cite{chen2024fine-alignment-acl-24,fse}, region-guided frameworks \cite{tanida-rgrg,tmi-2024-scene-graph-rrg}, warm starting \cite{nicolson-improving-cvt2distilgpt2}, patient-specific ``INDICATION" \cite{sei,mcl}, disease labels \cite{yang-m2kt}, and disease progression \cite{recap,2024-eccv-hergen}. However, these methods rely on single-image or fixed-view data, missing the comprehensive insights supplied by multi-view longitudinal data. To address this issue, we propose the MLRG method, which flexibly captures spatiotemporal features from multi-view longitudinal data and generates radiology reports based on patient-specific prior knowledge.

\begin{figure*}
    \centering
    \includegraphics[width=1\linewidth]{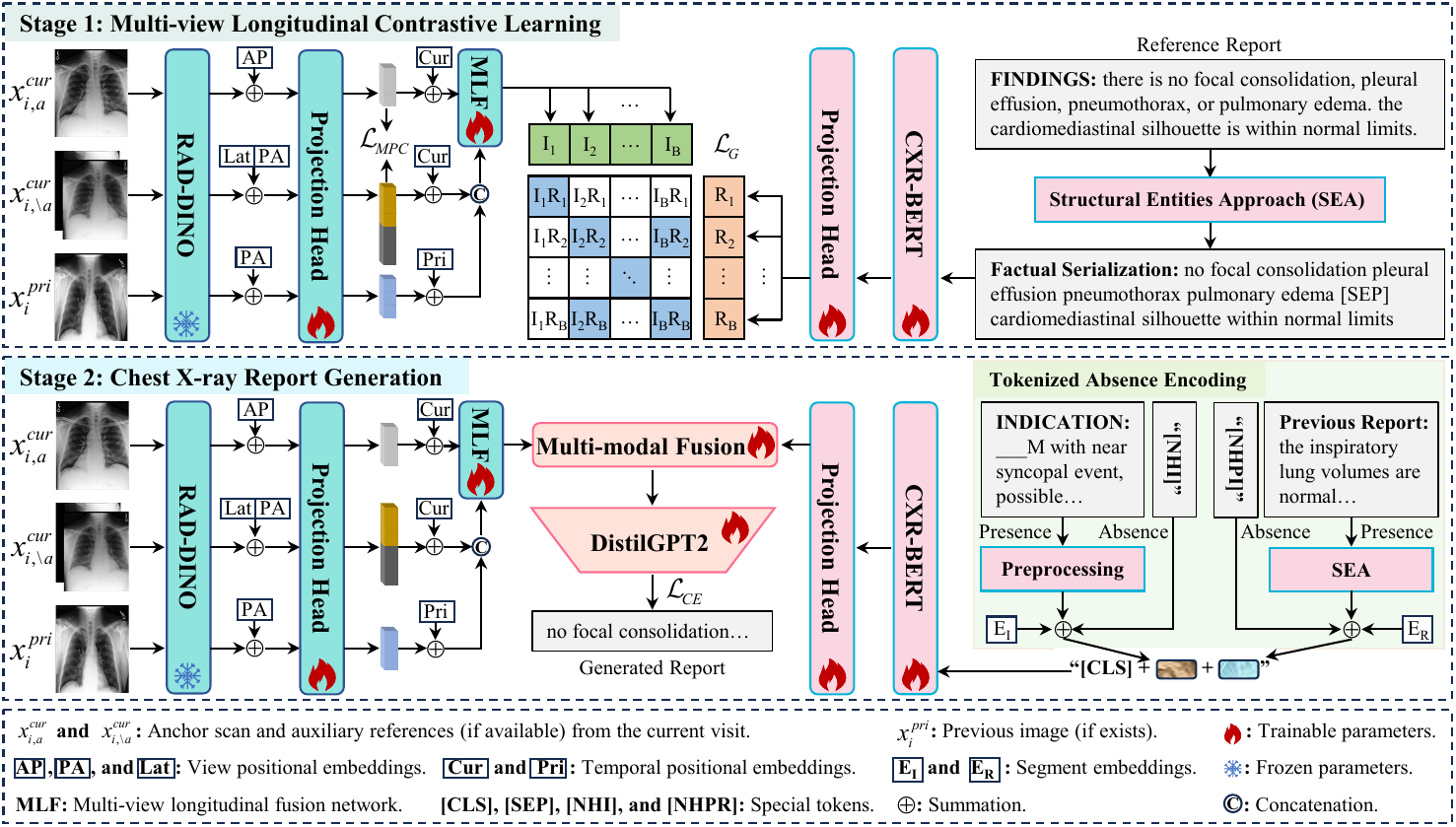}
    \caption{Overview of our proposed MLRG, including a vision encoder (RAD-DINO \cite{2024-rad-dino}), a text encoder (CXR-BERT \cite{2022-eccv-cxr-bert}), and a text generator (DistilGPT2 \cite{Sanh2019DistilBERTAD}). MLRG first learns visual features through multi-view longitudinal contrastive learning and then generates radiology reports based on patient-specific prior knowledge.}
    \label{fig:2}
\end{figure*}


\textbf{Medical vision-language models.} These models aim to learn generalized medical visual representations by maximizing agreements between image-report pairs. MGCA \cite{wang-mgca} presents multi-granularity cross-modal alignment, harnessing agreements at the instance, pathological region, and disease levels. KAD \cite{zhang-kad} enhances visual representation using established knowledge graphs. MedCLIP \cite{Wu-medclip} expands training sets by decoupling images and reports, reducing false negatives through semantic matching loss. BioViL-T \cite{2023-cvpr-biovit-2301} captures disease progression by analyzing longitudinal data. Despite notable improvements in tasks like medical image classification and image-text retrieval, the utilization of multi-view longitudinal data remains limited, constraining diagnostic accuracy. Therefore, we present a multi-view longitudinal contrastive learning approach that utilizes the inherent spatiotemporal information of radiology reports to guide the pre-training of visual and textual representations.


\textbf{Enhancing medical image analysis via multi-view data.} Multi-view learning \cite{a_survey_multi_view} empowers models to derive shared and complementary insights from multiple views of the same subject. FMVP \cite{tmm_mulview_2024-fmvp} treats single-view images and auxiliary inputs (i.e., disease labels and medical concepts) as multi-view data to assist the text generator in producing radiology reports. However, its reliance on additional annotated disease labels limits broader applicability. CXRMate \cite{nicolson2023-longitudinal-multiview} synthesizes previous reports based on previous images and integrates them with current multi-view images to produce final reports. However, this method overlooks subtle differences across views and can introduce additional noise from previous reports. In response, we propose the MLRG approach, which captures inter-view differences and leverages spatiotemporal information in reports to guide the pre-training, all without relying on additional manual labels.

\section{Method}


Figure \ref{fig:2} presents an overview of our proposed MLRG. In Stage 1, we introduce a multi-view longitudinal contrastive learning approach that leverages inherent spatiotemporal information from radiology reports to supervise the pre-training of visual and textual representations. In Stage 2, we propose a tokenized absence encoding technique to handle missing patient-specific prior knowledge, ensuring the generation of more coherent and accurate radiology reports based on available prior knowledge.

\subsection{Problem Formation}
Let ${\mathcal{D} _{tr}} = \{(x_i^{pri},y_i^{pri},{z_i},X_i^{cur},y_i^{cur})\} _{i=1}^{n}$ be the training set, where $n$ denotes the total number of visits. Each visit consists of a frontal previous image $x_i^{pri}$ (which may be absent), a previous report $y_i^{pri}$ (which may be absent), an ``INDICATION" ${z_i}$ (which may be absent), $m_i$ current images (views) $X_i^{cur}$, and a reference report $y_i^{cur}$. Notably, The number of current multi-view images $m_i$ may vary across visits. Our goal is to learn the function $F_{\theta}(\cdot)$ that maps $(x_i^{pri},y_i^{pri},{z_i},X_i^{cur})$ to $y_i^{cur}$ on the training set $\mathcal{D} _{tr}$, such that $F_\theta (x_i^{pri},y_i^{pri},{z_i},X_i^{cur})\to y_i^{cur}$. We then utilize the learned function $F_{\theta}(\cdot)$ to generate a radiology report based on current multi-view images, the previous image, and patient-specific prior knowledge (i.e., $y_i^{pri}$ and ${z_i}$).

\subsection{Multi-view Longitudinal Contrastive Learning}
\textbf{Visual features extraction.} We employ RAD-DINO \cite{2024-rad-dino}, a vision transformer model \cite{2021-vit} trained solely on chest X-rays using DINOv2 \cite{2024-dinov2}, as the vision encoder. The feature maps from the last hidden state are treated as visual features ${\boldsymbol{V}} \in {\mathbb{R}^{M \times p \times d_1}}$, where $M = \sum\nolimits_{i = 1}^B {{m_i}}$ denotes the total number of images in the mini-batch. Here, $B$, $p$, and $d_1$ represent the batch size, the number of patches, and the feature dimension, respectively.

\textbf{Textual features extraction.} Inspired by FSE \cite{fse}, we first adopt the structural entities approach \cite{fse} to extract factual serialization, which consists exclusively of clinical descriptions from radiology reports, as shown in Figure \ref{fig:2}. This method enables the model to concentrate on alignment between images and factual serialization. We then consider CXR-BERT \cite{2022-eccv-cxr-bert}, a language model tailored for chest X-rays, as the text encoder. This is followed by a simple projection head that generates textual features ${\boldsymbol{R}}\in {\mathbb{R}^{B \times t \times d}}$, where $t$ denote the number of tokens, and $d$ is the hidden size.

\textbf{Multi-positive contrastive learning between current multi-view images to improve the consistency of visual features.} In clinical practice, radiologists often select some representative images as primary references, with other images as auxiliary support. Therefore, we treat each image from current multi-view images $X_i^{cur}$ as an anchor scan $x_{i, a}^{cur}$ while considering the remaining images as auxiliary references $x_{i,\backslash a}^{cur}=\left \{ x_{i,j}^{cur}|j \neq a, x_{i,j}^{cur} \in X_i^{cur} \right \} $, where $a \in [1,m_i]$. To identify differences among views, we incorporate learnable view positional embeddings ${\boldsymbol{E}}_v \in {\mathbb{R}^{M \times 1 \times d_1}}$ into visual features. This is followed by a simple projection head $P_v(\cdot )$ that maps the features to a specific dimension $d$. These processes are formulated as follows:
\begin{align}
{\boldsymbol{V}} = P_v\left ( {\boldsymbol{V}} + {\boldsymbol{E}}_v \right ) \in {\mathbb{R}^{M \times p \times d}}.
\end{align}
To enhance the consistency of visual features, we employ multi-positive contrastive learning \cite{NEURIPS2023_stablerep} to maximize the similarity between images from the same visit while minimizing the similarity to images from different visits. Specifically, we first exclude visits with only one image, as they do not provide positive pairs (Notably, these visits are still used for subsequent cross-modal alignment). Following this, we calculate the predicted categorical distribution ${\boldsymbol{q}} \in {\mathbb{R}^{K \times (K-1)}}$ to estimate the similarity between images:
\begin{align}
{{\boldsymbol q}_i} = \frac{{\exp \left( {{{{{\boldsymbol v}_i} \cdot {{\boldsymbol v}_{\backslash i}^T}} \mathord{\left/ \right.
 } {{\tau _1}}}} \right)}}{{\sum\nolimits_{j = 1,j \ne i}^M {\exp \left( {{{{{\boldsymbol v}_i} \cdot {{\boldsymbol v}_j^T}} \mathord{\left/ \right.
 } {{\tau _1}}}} \right)} }}, s.t. m_i \ne 1,
\end{align}
where $K = \sum\nolimits_{i = 1,m_i \ne 1}^M {{m_i}}$ represents the total number of multi-view images in the mini-batch, and $\tau_1 \in {\mathbb{R}^{+}}$ is a temperature parameter. ${{\boldsymbol v}_i} \in {\mathbb{R}^{1 \times d}}$ refers to the global visual feature of the $i^{th}$ image, while ${{\boldsymbol v}_{\backslash i}} \in {\mathbb{R}^{(K-1) \times d}}$ denotes global visual features of all multi-view images except the $i^{th}$ image. Next, we compute the ground-truth categorical distribution ${\boldsymbol{p}} \in {\mathbb{R}^{K \times (K-1)}}$ by assigning the same labels to images from the same visit, formulated as:
\begin{align}
{\boldsymbol{p}_i} = \frac{{{\mathbb{I}_{match}}\left( {{{\boldsymbol v}_{i}},{{{\boldsymbol v}}_{\backslash i}}} \right)}}{{\sum\nolimits_{j = 1,m_j \ne 1}^{M} {{\mathbb{I}_{match}}\left( {{{{\boldsymbol v}}_{i}},{{\boldsymbol v}}_{j}} \right)} }},
\end{align}
where \({\mathbb{I}_{match}}\left( {\cdot, \cdot} \right)\) is an indicator function that determines whether two visual features originate from the same visit. Although the number of current views $m_i$ may vary across visits, the different number of non-zero elements in \({\boldsymbol{p}_i}\) account for this variability. Finally, the multi-positive contrastive (MPC) loss is calculated using the cross-entropy between $\boldsymbol{q}$ and $\boldsymbol{p}$, represented as:
\begin{align}
{{{\cal L}}_{MPC}} =  - \frac{1}{K}\sum\limits_{i = 1,m_i \ne 1}^{M} {{{\boldsymbol{p}}_i}\log {{\boldsymbol{q}}_i}}.
\end{align}

\begin{figure}
    \centering
    \includegraphics[width=1\linewidth]{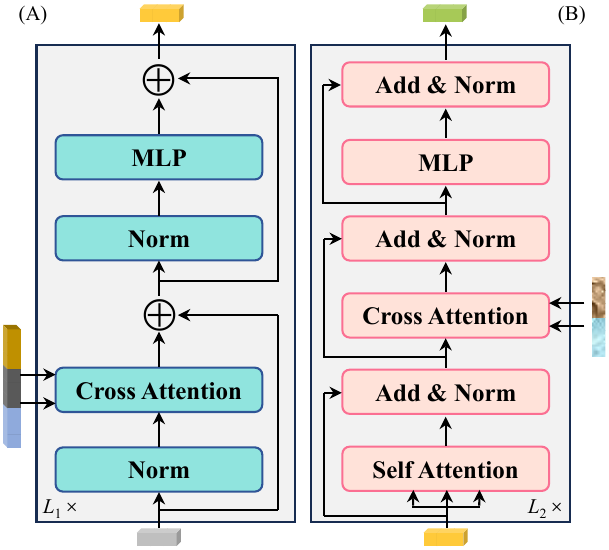}
    \caption{(A) represents the multi-view longitudinal fusion (MLF) network. (B) denotes the multi-modal fusion network.}
    \label{fig:3}
\end{figure}

\begin{table*}
\centering
\setlength{\tabcolsep}{0.9mm}
\begin{tabular}{cccccccccccccccc} 
\toprule
\multirow{2}{*}{\textbf{Split} ~} & \multicolumn{5}{c}{\textbf{MIMIC-CXR}} & \multicolumn{5}{c}{\textbf{MIMIC-ABN}} & \multicolumn{5}{c}{\textbf{Two-view CXR}} \\ 
\cmidrule(lr){2-6}\cmidrule(r){7-11}\cmidrule(lr){12-16}
 & \textbf{\#Img} & \textbf{\#Rep} & \textbf{\%Ind} & \textbf{\%PI} & \textbf{\%PR} & \textbf{\#Img} & \textbf{\#Rep} & \textbf{\%Ind} & \textbf{\%PI} & \textbf{\%PR} & \textbf{\#Img} & \textbf{\#Rep} & \textbf{\%Ind} & \textbf{\%PI} & \textbf{\%PR} \\ 
\midrule
Train & 239,998 & 150,957 & 66.4 & 60.5 & 60.5 & 69,641 & 34,763 & 64.6 & 52.7 & 52.7 & 181,312 & 90,656 & 67.1 & 51.6 & 51.6 \\
Val & 2,113 & 1,182 & 65.4 & 61.3 & 0.0 & 586 & 263 & 62.7 & 50.6 & 0.0 & 1,778 & 889 & 72.1 & 36.2 & 0.0 \\
Test & 3,852 & 2,343 & 57.3 & 87.7 & 0.0 & 844 & 378 & 56.3 & 89.2 & 0.0 & 3,000 & 1500 & 68.9 & 55.8 & 0.0 \\
\bottomrule
\end{tabular}
\caption{Statistics of the three datasets for the training, validation, and test sets. ``\#Img" and ``\#Rep" denote the number of images and reports, while ``\%Ind", ``\%PI", and ``\%PR" represent the ratios of ``INDICATION", ``previous image", and ``previous report", respectively.}
\label{table:0}
\end{table*}

\textbf{Multi-view longitudinal fusion network.} Due to the varying number of current multi-view images and the absence of previous images for some patients, integrating this information flexibly presents certain challenges. To address this issue, we design the multi-view longitudinal fusion (MLF) network, as illustrated in Figure \ref{fig:3}(A). We select the most recent previous visit to model temporal information, as it typically holds the highest reference value. To distinguish different time points, we integrate temporal positional embeddings into visual features, as depicted in Figure \ref{fig:2}. Subsequently, the spatiotemporal features ${\boldsymbol V}^{st}=\left \{ {\boldsymbol v}^{st}_1, {\boldsymbol v}^{st}_2,\dots ,{\boldsymbol v}^{st}_B\right \} $ are extracted using the MLF network:
\begin{align}
{\boldsymbol v}^{st}_i=\text{MLF} ({\boldsymbol v}_{i,a}^{cur},[{\boldsymbol v}_{i, \backslash a}^{cur},{\boldsymbol v}^{pri}_i]) \in {\mathbb{R}^{p \times d}},
\end{align}
where the anchor scan ${\boldsymbol v}_{i,a}^{cur}$ functions as the query, and the concatenation of auxiliary references and previous image, $[{\boldsymbol v}_{i, \backslash a}^{cur},{\boldsymbol v}^{pri}_i]$, serves as the key and value. Although the number of images for the current visit varies and some patients may lack previous images, we process one sample at a time, allowing for flexible adaptation to these changes.

\textbf{Instance-wise cross-modal alignment.} Radiology reports not only describe the current visit's condition but may also include comparisons with the patient's medical history. Therefore, relying solely on the current multi-view images and corresponding reports for cross-modal alignment could lead to model hallucinations. To address this, We first extract spatiotemporal features from multi-view longitudinal data using the multi-view longitudinal fusion network (see Figure \ref{fig:3}(A)). Subsequently, we utilize the inherent spatiotemporal information from radiology reports to supervise the pre-training of visual and textual representations. Inspired by CLIP \cite{radford-learning-clip} and MGCA \cite{wang-mgca}, we employ instance-wise cross-modal alignment to learn uni-modal representations. Specifically, we compute the image-to-text predicted categorical distribution ${\boldsymbol{q}}^{v2r} \in {\mathbb{R}^{B \times B}}$, defined as:
\begin{align}
{{\boldsymbol q}^{v2r}} = \frac{{\exp \left( {{{{{\boldsymbol {\bar V} }} \cdot {{\boldsymbol {\bar R}}^T}} \mathord{\left/ \right.
 } {{\tau _2}}}} \right)}}{{\sum\nolimits_{j = 1}^B {\exp \left( {{{{{\boldsymbol {\bar V}}} \cdot {{{\boldsymbol {\bar R}}_j^T}}} \mathord{\left/ \right.
 } {{\tau _2}}}} \right)} }}, 
\end{align}
where ${\boldsymbol {\bar V}}$ and ${\boldsymbol {\bar R}}$ denote global features of spatiotemporal features ${\boldsymbol {V}^{st}}$ and textual features ${\boldsymbol {R}}$, respectively. Similarly, we can also obtain the symmetric text-to-image predicted categorical distribution ${\boldsymbol{q}}^{r2v}$. Given that radiology report content may be consistent across visits, we consider both spatiotemporal and textual features from the same visit as positive pairs, as well as those from different visits with identical reports. Consequently, the image-to-text ground-truth categorical distribution ${\boldsymbol{p}}^{v2r} \in {\mathbb{R}^{B \times B}}$ is defined as:
\begin{align}
{\boldsymbol{p}_{i,j}^{g}} = \frac{{{\mathbb{I}_{identical}}\left( {{y_{i}^{cur}}, {y_{j}^{cur}}} \right)}}{{\sum\nolimits_{k = 1}^{B} {{\mathbb{I}_{identical}}\left( {{y_{i}^{cur}},{y_{k}^{cur}}} \right)} }},
\end{align}
where ${{\mathbb{I}_{identical}}\left( {\cdot,\cdot} \right)}$ denotes an indicator function that determines whether two radiology reports are identical. Finally, the cross-modal alignment loss is defined as:
\begin{align}
{{{\cal L}}_{G}} =  - \frac{1}{B}\sum\limits_{i = 1}^B {\left( {{\boldsymbol{p}}_i^{g}\log {\boldsymbol{q}}_i^{v2r} + {\boldsymbol{p}}_i^{g}\log {\boldsymbol{q}}_i^{r2v}} \right)}.
\end{align}

\textbf{Overall objective in Stage 1.} We train our MLRG by jointly optimizing ${\cal L}_{MPC}$ and ${\cal L}_{G}$, 
formulated as:
\begin{align}
{{{\cal L}}_{pretrain}} = {{\cal L}}_{MPC} + {{\cal L}}_{G}.
\end{align}


\begin{table*}
\centering
\setlength{\tabcolsep}{1.15mm}
\begin{tabular}{c|ccccccccccccc} 
\toprule
\multicolumn{1}{c}{\multirow{2}{*}{\textbf{Dataset}}} & \multirow{2}{*}{\textbf{Input}} & \multirow{2}{*}{\textbf{Method}} & \multirow{2}{*}{\textbf{Venue}} & \multicolumn{6}{c}{\textbf{NLG Metrics} $\uparrow$} & \multicolumn{4}{c}{\textbf{CE Metrics} $\uparrow$} \\ 
\cmidrule(r){5-10}\cmidrule(r){11-14}
\multicolumn{1}{c}{} &  &  &  & \textbf{B-1} & \textbf{B-2} & \textbf{B-3} & \textbf{B-4} & \textbf{MTR} & \textbf{R-L} & \textbf{RG} & \textbf{P} & \textbf{R} & \textbf{F1} \\ 
\midrule
\multirow{14}{*}{M-CXR} & SI & SA \cite{yan2023style} & EMNLP'23 & - & 0.184 & - & - & - & - & 0.228 & - & - & 0.394 \\
 & SI & MET \cite{wang2023metransformer} & CVPR'23 & 0.386 & 0.250 & 0.169 & 0.124 & 0.152 & 0.291 & - & 0.364 & 0.309 & 0.311 \\
 & SI & KiUT \cite{huang-kiut} & CVPR'23 & 0.393 & 0.243 & 0.159 & 0.113 & 0.160 & 0.285 & - & 0.371 & 0.318 & 0.321 \\
 & SI & CoFE \cite{cofe-eccv-24} & ECCV'24 & - & - & - & 0.125 & \textbf{0.176} & \underline{0.304} &  & 0.489 & 0.370 & 0.405 \\
 & SI & MAN \cite{shen2024automatic_aaai} & AAAI'24 & 0.396 & 0.244 & 0.162 & 0.115 & 0.151 & 0.274 & - & 0.411 & 0.398 & 0.389 \\
 & SI & B-LLM \cite{aaai-liu2024bootstrapping-llm} & AAAI'24 & \underline{0.402} & \underline{0.262} & \underline{0.180} & 0.128 & \underline{0.175} & 0.291 & - & 0.465 & \textbf{0.482} & \underline{0.473} \\
 & SI & DCG \cite{liang2024divide-acmmm-24} & ACMMM'24 & 0.397 & 0.258 & 0.166 & 0.126 & 0.162 & 0.295 & - & 0.441 & 0.414 & 0.404 \\
 & SI & Med-LLM \cite{liu2024in-context-acmmm} & ACMMM'24 & - & - & - & 0.128 & 0.161 & 0.289 & - & 0.412 & 0.373 & 0.395 \\
 & SI+Ind & SEI \cite{sei} & MICCAI'24 & 0.382 & 0.247 & 0.177 & \underline{0.135} & 0.158 & 0.299 & \underline{0.249} & \underline{0.523} & 0.410 & 0.460 \\
 & MVD & FMVP \cite{tmm_mulview_2024-fmvp} & TMM'23 & 0.389 & 0.236 & 0.156 & 0.108 & 0.150 & 0.284 & - & 0.332 & 0.383 & 0.336 \\
 & Long & HERGen \cite{2024-eccv-hergen} & ECCV'24 & 0.395 & 0.248 & 0.169 & 0.122 & 0.156 & 0.285 & - & - & - & - \\
 & MVL & CXRMate \cite{nicolson2023-longitudinal-multiview} & arXiv'23 & 0.361 & 0.223 & 0.150 & 0.108 & 0.159 & 0.263 & 0.238 & 0.495 & 0.367 & 0.422 \\
 & MVL & \textbf{MLRG(Ours)} & - & \textbf{0.411} & \textbf{0.277} & \textbf{0.204} & \textbf{0.158} & \textbf{0.176} & \textbf{0.320} & \textbf{0.291} & \textbf{0.549} & \underline{0.468} & \textbf{0.505} \\ 
\cline{2-14}
 & - & $\Delta$ (\%) $\uparrow$ & - & +0.9 & +1.5 & +2.4 & +2.3 & +0.1 & +1.6 & +4.2 & +2.6 & -1.4 & +3.2 \\ 
\cmidrule(r){1-14}
\multirow{5}{*}{M-ABN} & SI & R2Gen$^\flat$ \cite{chen-etal-2020-generating} & EMNLP'20 & 0.253 & 0.144 & 0.092 & 0.063 & 0.106 & 0.229 & 0.179 & 0.444 & 0.425 & 0.434 \\
 & SI & CMN$^\flat$ \cite{chen-etal-2021-cross-modal} & ACL'21 & 0.256 & 0.147 & 0.095 & 0.066 & 0.110 & 0.230 & 0.183 & \underline{0.466} & \underline{0.454} & \underline{0.460} \\
 & SI+Ind & SEI$^\flat$ \cite{sei} & MICCAI'24 & \underline{0.267} & \underline{0.157} & \underline{0.104} & \underline{0.073} & \underline{0.114} & \underline{0.231} & \underline{0.191} & \underline{0.466} & 0.408 & 0.435 \\
 & MVL & \textbf{MLRG(Ours)} & - & \textbf{0.332} & \textbf{0.199} & \textbf{0.132} & \textbf{0.094} & \textbf{0.136} & \textbf{0.248} & \textbf{0.219} & \textbf{0.513} & \textbf{0.517} & \textbf{0.515} \\ 
\cline{2-14}
 & - & $\Delta$ (\%) $\uparrow$ & - & +6.5 & +4.2 & +2.8 & +2.1 & +2.2 & +1.7 & +2.8 & +4.7 & +6.3 & +5.5 \\ 
\cmidrule(r){1-14}
\multirow{5}{*}{T-CXR} & DV & R2Gen$^\flat$ \cite{chen-etal-2020-generating} & EMNLP'20 & 0.346 & 0.219 & 0.153 & 0.113 & 0.141 & 0.302 & 0.267 & 0.478 & 0.329 & 0.390 \\
 & DV & CMN$^\flat$ \cite{chen-etal-2021-cross-modal} & ACL'21 & 0.387 & 0.241 & 0.166 & 0.122 & 0.151 & 0.310 & 0.268 & 0.496 & 0.336 & 0.401 \\
 & DV+Ind & SEI$^\flat$ \cite{sei} & MICCAI'24 & \underline{0.409} & \underline{0.263} & \underline{0.186} & \underline{0.140} & \underline{0.168} & \underline{0.320} & \underline{0.301} & \underline{0.522} & \underline{0.447} & \underline{0.481} \\
 & MVL & \textbf{MLRG(Ours)} & - & \textbf{0.417} & \textbf{0.276} & \textbf{0.200} & \textbf{0.154} & \textbf{0.178} & \textbf{0.331} & \textbf{0.328} & \textbf{0.532} & \textbf{0.474} & \textbf{0.501} \\ 
\cline{2-14}
 & - & $\Delta$ (\%) $\uparrow$ & - & +0.8 & +1.3 & +1.4 & +1.4 & +1.0 & +1.1 & +2.7 & +1.0 & +2.7 & +2.0  \\
\bottomrule
\end{tabular}
\caption{Comparison with SOTA methods on MIMIC-CXR (M-CXR), MIMIC-ABN (M-ABN), and Two-view CXR (T-CXR) datasets. \(\Delta\) denotes the performance difference between MLRG and the best peer methods. \({\flat}\) signifies results reproduced using official codes, while other results are sourced from original publications. The best and second-best values are emphasized in \textbf{bold} and \underline{underlined}, respectively.}
\label{table:1}
\end{table*}

\subsection{Chest X-ray Report Generation}
\textbf{Integrating patient-specific prior knowledge into the text generator.} Radiologists commonly refer to patient-specific prior knowledge, such as the ``INDICATION" (which outlines the visit reasons or symptoms) and the ``previous report" (which provides the patient's medical history), when drafting radiology reports. However, these data may be absent for some patients due to de-identification or incomplete records. To combat this issue, we propose a tokenized absence encoding technique to handle missing patient-specific prior knowledge, as shown in Figure \ref{fig:2}. Specifically,  for missing ``INDICATION" and ``previous report", we utilize special tokens, ``[NHI]" and ``[NHPR]", to simulate their presence. For existing  ``INDICATION", we apply a preprocessing strategy from SEI \cite{sei} to remove de-identification noise (e.g., \textit{-year-old}, \_\_\_, and \textbackslash\textbackslash). For available ``previous report", we employ the structural entities approach \cite{fse} to extract factual serialization, enabling the model to focus on clinically relevant details. We then combine the cleaned ``INDICATION" with factual serialization extracted from the ``previous report" into a cohesive paragraph, supplemented by segment embeddings to help the model distinguish between different sentence meanings.  Finally, we utilize the multi-modal fusion network \cite{chen-ptunifier} (illustrated in Figure \ref{fig:3}(B)) to flexibly integrate patient-specific prior knowledge into the spatiotemporal features ${\boldsymbol V}^{st}$, allowing the text generator to generate more accurate radiology reports based on available prior knowledge.

\textbf{Report generation.} We start by initializing the projection heads, MLF network, and CXR-BERT using the model trained in Stage 1. Following prestigious works \cite{nicolson-improving-cvt2distilgpt2,2024-eccv-hergen}, we treat the DistilGPT2 \cite{Sanh2019DistilBERTAD}, initialized by CGPT2 \cite{nicolson-improving-cvt2distilgpt2}, as the text generator. We then minimize the cross-entropy loss ${\cal L}_{CE}$ to ensure that the generated reports closely align with the reference reports.


\section{Experiments}

\subsection{Experimental Settings}
\textbf{Datasets.} 1) \textbf{MIMIC-CXR} \cite{johnson-mimic-cxr-jpg} is a large-scale, publicly available dataset comprising paired chest X-rays and radiology reports. Each pair contains a varying number of images compared to others, and all pairs for a patient are organized chronologically, facilitating the construction of multi-view longitudinal data. 2) \textbf{MIMIC-ABN} \cite{mimic-abn-ori} is a subset of MIMIC-CXR, focusing solely on radiology reports that describe abnormal clinical findings. 3) \textbf{Two-view CXR} \cite{mcl} aggregates visits with two current images from both MIMIC-CXR and IU X-ray \cite{demner2016preparing-iu-xray}. Notably, as the IU X-ray does not include previous visits, both previous images and reports are unavailable. We adhere to official splits for these datasets and summarize the sample counts for the training, validation, and test set in Table \ref{table:0}. In line with \cite{chen-etal-2020-generating,MMTN-aaai-2023,tanida-rgrg,sei,li-dcl}, we treat the ``FINDINGS" section in radiology reports as the reference reports.

\begin{table*}
\centering
\setlength{\tabcolsep}{1.8mm}
\begin{tabular}{cccccccccccccccc} 
\toprule
\multirow{2}{*}{\textbf{Model}} & \multirow{2}{*}{\textbf{M/S}} & \multirow{2}{*}{\textbf{F/R}} & \multirow{2}{*}{\textbf{PI}} & \multicolumn{2}{c}{\textbf{Stage 1}} & \multicolumn{2}{c}{\textbf{Stage 2}} & \multicolumn{6}{c}{\textbf{NLG metrics} $\uparrow$} & \multicolumn{2}{c}{\textbf{CE metrics} $\uparrow$} \\ 
\cmidrule(lr){5-6}\cmidrule(lr){7-8}\cmidrule(lr){9-14}\cmidrule(r){15-16}
 &  &  &  & ${\cal{L}}_{G}$ & ${\cal{L}}_{MPC}$ & \textbf{Ind} & \textbf{PR} & \textbf{B-1} & \textbf{B-2} & \textbf{B-3} & \textbf{B-4} & \textbf{MTR} & \textbf{R-L} & \textbf{RG} & \textbf{F1} \\ 
\midrule
(a) & M & F & \ding{51} & \ding{55} & \ding{55} & \ding{51} & \ding{51} & 0.346 & 0.235 & 0.173 & 0.136 & 0.154 & 0.305 & 0.258 & 0.373 \\
(b) & M & F & \ding{51} & \ding{51} & \ding{51} & \ding{55} & \ding{55} & 0.385 & 0.239 & 0.162 & 0.118 & 0.155 & 0.283 & 0.238 & 0.479 \\
(c) & M & F & \ding{55} & \ding{51} & \ding{55} & \ding{51} & \ding{51} & 0.384 & 0.257 & 0.188 & 0.146 & 0.165 & 0.310 & 0.267 & 0.455 \\
(d) & M & F & \ding{55} & \ding{51} & \ding{51} & \ding{51} & \ding{51} & 0.395 & 0.257 & 0.183 & 0.138 & 0.167 & 0.302 &0.278~ & 0.503 \\
(e) & M & F & \ding{51} & \ding{51} & \ding{51} & \ding{51} & \ding{55} & 0.392 & 0.265 & 0.195 & 0.153 & 0.171 & 0.316 & 0.281 & 0.476 \\
(f) & M & F & \ding{51} & \ding{51} & \ding{51} & \ding{55} & \ding{51} & 0.387 & 0.240 & 0.163 & 0.118 & 0.156 & 0.281 & 0.243 & 0.484 \\
(g) & M & R & \ding{51} & \ding{51} & \ding{51} & \ding{51} & \ding{51} & 0.403 & 0.267 & 0.193 & 0.148 & 0.172 & 0.309 & 0.287 & \textbf{0.510} \\
(h) & S & F & \ding{51} & \ding{51} & \ding{55} & \ding{51} & \ding{51} & 0.400 & 0.269 & 0.196 & 0.151 & 0.173 & 0.314 & 0.289 & 0.498 \\
\textbf{MLRG} & M & F & \ding{51} & \ding{51} & \ding{51} & \ding{51} & \ding{51} & \textbf{0.411} & \textbf{0.277} & \textbf{0.204} & \textbf{0.158} & \textbf{0.176} & \textbf{0.320} & \textbf{0.291} & 0.505 \\
\bottomrule
\end{tabular}
\caption{Ablation study on the MIMIC-CXR dataset. ``M/S" refers to methods that utilize current \textbf{M}ulti-view images or \textbf{S}ingle images as input. ``F/R" indicates alignment based on either \textbf{F}actual serialization or \textbf{R}eport. ``PI", ``PR", and ``Ind" represent \textbf{P}revious \textbf{I}mages, \textbf{P}revious \textbf{R}eports, and ``INDICATION", respectively. The best values are emphasized in \textbf{bold}.}
\label{table:2}
\end{table*}

\textbf{Evaluation metrics.} Following prior works \cite{chen-etal-2020-generating,Wang_2021_CVPR_region,wang2023metransformer,2024-eccv-hergen,Bu_Instance-level-2024_CVPR}, we evaluate the effectiveness of our MLRG using both natural language generation (NLG) and clinical efficacy (CE) metrics. NLG metrics, which assess the linguistic similarities between generated and reference reports, include BLEU-n (B-n), METEOR (MTR), and ROUGE-L (R-L). For CE metrics, we utilize CheXpert \cite{irvin-chexpert} to label the generated reports with 14 observations (see Appendix Table \ref{atable:1}) and compute the micro-average Precision (P), Recall (R), and F1 score (F1) based on ground truths. CE metrics also include F1 RadGraph (RG) \cite{jain-radgraph}, which evaluates the overlap of clinical entities and their relations, aligning more closely with radiologists than B-3 and F1 metrics \cite{yu2023evaluating}. All metrics are computed using official libraries \cite{chen2015microsoft-coco-caption,jain-radgraph,Smit2020_chexbert}, with higher values indicating better performance.

\textbf{Implementation details.} Both $\tau_1$ and $\tau_2$ are set to 0.5. The number of blocks $L_1$ and $L_2$ in Figure \ref{fig:3} are set to 3 and 1, respectively. Each dataset is configured to generate a maximum of 100 tokens. We identify the best model as the one with the highest sum of BLEU-4, F1 RadGraph, and F1 score on the validation set, and we report its performance on the test set. Additional details about epochs, learning rates, and other settings can be found in the Appendix \ref{implementation-details}.


\begin{table}
\centering
\setlength{\tabcolsep}{1.4mm}
\begin{tabular}{ccccccc} 
\toprule
\textbf{Setting} & \textbf{\%} & \textbf{B-2 $\uparrow$} & \textbf{B-4 $\uparrow$} & \textbf{MTR $\uparrow$} & \textbf{R-L $\uparrow$} & \textbf{RG $\uparrow$} \\ 
\midrule
w/ Ind & 57.8 & \textbf{0.295} & \textbf{0.174} & \textbf{0.184} & \textbf{0.332} & \textbf{0.318} \\
w/o Ind & 42.2 & 0.253 & 0.137 & 0.166 & 0.302 & 0.254 \\
w/ MV & 70.7 & \textbf{0.282} & \textbf{0.161} & \textbf{0.179} & \textbf{0.323} & \textbf{0.301} \\
w/o MV & 29.3 & 0.264 & 0.150 & 0.171 & 0.310 & 0.266 \\
w/ MVL & 61.4 & \textbf{0.282} & \textbf{0.160} & \textbf{0.178} & \textbf{0.322} & \textbf{0.300} \\
w/o MVL & 38.6 & 0.270 & 0.155 & 0.174 & 0.316 & 0.276 \\
\bottomrule
\end{tabular}
\caption{Breakdown of MLRG's metrics on the MIMIC-CXR test set, categorized by (a) inclusion of indications (Ind), (b) inclusion of current multi-view images (MV), (c) inclusion of multi-view longitudinal data (MVL).}
\label{table:3}
\end{table}

\subsection{Main Results}
We compare our MLRG with 14 state-of-the-art methods: R2Gen \cite{chen-etal-2020-generating}, CMN \cite{chen-etal-2021-cross-modal}, SA \cite{yan2023style}, MET \cite{wang2023metransformer}, KiUT \cite{huang-kiut}, CoFE \cite{cofe-eccv-24}, MAN \cite{shen2024automatic_aaai}, B-LLM \cite{aaai-liu2024bootstrapping-llm}, DCG \cite{liang2024divide-acmmm-24}, Med-LLM \cite{liu2024in-context-acmmm}, SEI \cite{sei}, FVMP \cite{tmm_mulview_2024-fmvp}, HERGen \cite{2024-eccv-hergen}, and CXRMate \cite{nicolson2023-longitudinal-multiview}. Results are presented in Table \ref{table:1}, where ``SI", ``Ind", ``MVD", ``Long", ``MVL", and ``DV" represent different input types: single images, ``INDICATION", multi-view data, longitudinal data, multi-view longitudinal data, and dual views, respectively. We observe that our MLRG achieves SOTA performance across most metrics, with particular strength in B-4, RG, and F1. This suggests that MLRG excels in generating both coherent and accurate radiology reports. Although MLRG shows slightly lower Recall than B-LLM \cite{aaai-liu2024bootstrapping-llm}, its F1 and other metrics are significantly better. In Appendix Table \ref{atable:3}, we also show our MLRG's ability to generate ``FINDINGS'' and ``IMPRESSION'' sections.

\subsection{Ablation Study}
Table \ref{table:2} presents an ablation study on the MIMIC-CXR \cite{johnson-mimic-cxr-jpg} dataset, analyzing the effect of different components on model performance.

\textbf{Effect of multi-view longitudinal contrastive learning (Stage 1).} In Table \ref{table:2}, (a) represents a report generation scheme based solely on patient-specific prior knowledge, excluding Stage 1. Results reveal that our MLRG significantly exceeds (a), highlighting the critical role of multi-view longitudinal contrastive learning in enhancing the accuracy and coherence of generated reports. Additionally, we observe that both  ${\cal{L}}_{G}$ ((c) vs. (a)) and ${\cal{L}}_{MPC}$ ((d) vs. (c)) have a positive impact on model performance.

\textbf{Effect of patient-specific prior knowledge (Stage 2).} As shown in Table \ref{table:2}, MLRG significantly outperforms (b), which lacks patient-specific prior knowledge, emphasizing the importance of incorporating such knowledge to improve the coherence and clinical accuracy of generated reports. Moreover, the independent integration of ``INDICATION" ((e) vs. (b)) and ``previous report" ((f) vs. (b)) contributes positively to model performance.

\textbf{Effect of current multi-view images.} Table \ref{table:2} demonstrates that generating reports using current multi-view images outperforms those derived from single images (MLRG vs. (h)), highlighting the effectiveness of multi-view images in modeling the current disease conditions. 

\textbf{Effect of previous images.} As shown in Table \ref{table:2}, MLRG shows a clear advantage over (d), indicating that MLF network (see Figure \ref{fig:3}) effectively integrates previous images. This capability allows the model to track disease progression, thereby generating more clinically accurate reports.

\begin{figure*}
    \centering
    \includegraphics[width=1\linewidth]{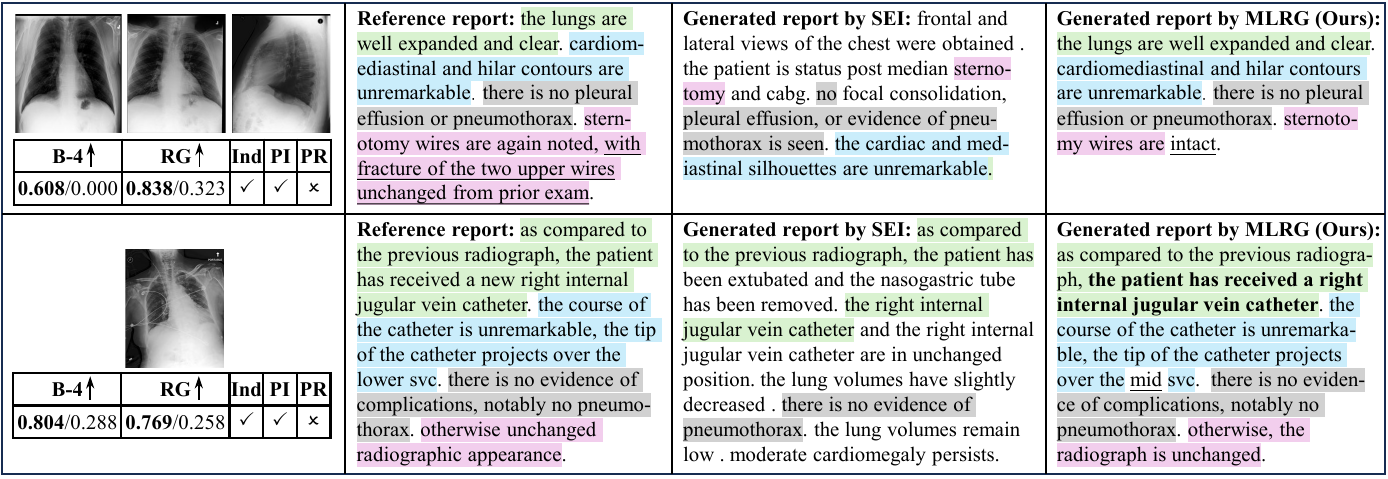}
    \caption{Generated reports examples on the MIMIC-CXR test set. Each ``A/B" cell refers to ``MLRG/SEI". Sentences in the reference report are highlighted in unique colors to clarify alignment with descriptions in the generated reports. Matching content in generated reports is shown in the same color, while correct temporal descriptions and failure descriptions of our MLRG are in \textbf{bold} and \underline{underlined}.}
    \label{fig:4}
\end{figure*}

\begin{figure}
    \centering
    \includegraphics[width=1\linewidth]{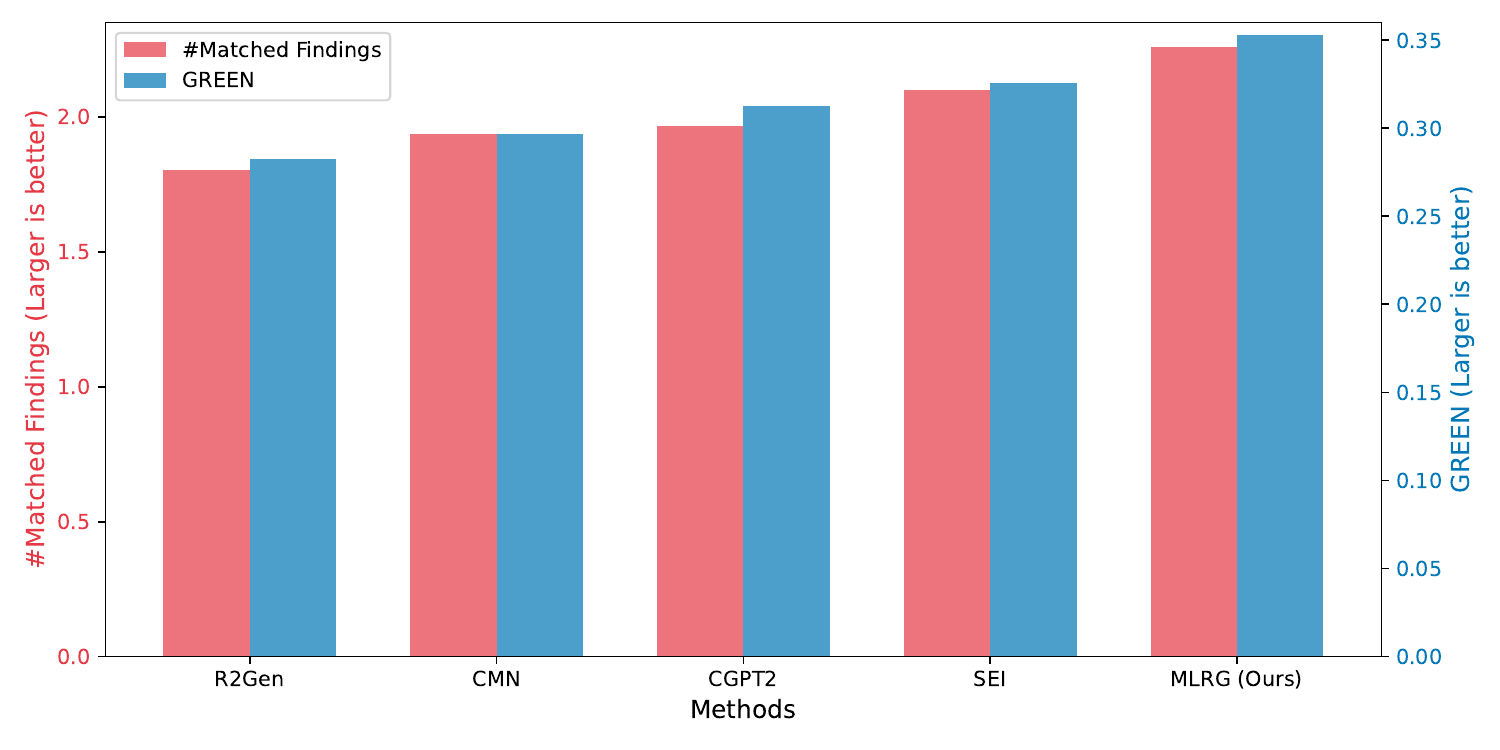}
    \caption{Comparison with baselines on MIMIC-CXR using LLMs. ``\#Matched Findings" denotes the number of matched findings between generated and reference reports.}
    \label{fig:5}
\end{figure}

\subsection{Case Study}
\textbf{Model benefits from current multi-view images, multi-view longitudinal data, and ``INDICATION".} Table \ref{table:3} compares performance on test subsets with and without these data (multi-view images, multi-view longitudinal data, and indications). We observe that including these data significantly improves NLG and RG metrics. This suggests that the multi-view longitudinal contrastive learning method effectively integrates current multi-view images and multi-view longitudinal data, capturing semantically rich visual representations. Moreover, our multi-modal fusion network effectively leverages ``INDICATION" to help the text generator produce more accurate radiology reports.

\textbf{Clinical accuracy of 14 observations.} Appendix Tables \ref{atable:1} and \ref{atable:2} show the clinical accuracy of 14 observations labeled by CheXpert \cite{irvin-chexpert} across three datasets. Results indicate that our MLRG outperforms SEI \cite{sei} on most observations. Although not specifically tailored for imbalanced observations, MLRG still slightly exceeds the baseline on challenging observations like \textit{Pneumothorax} and \textit{Fracture}.

\textbf{Qualitative analysis.} Figure \ref{fig:4} presents examples of generated reports using SEI \cite{sei} and our MLRG, with additional examples in Appendix Figure \ref{afig:1}. A greater number of colors in the generated report indicates broader coverage of clinical findings, while a longer color bar reflects more accurate and detailed descriptions of specific findings. Results indicate that 1) Our MLRG provides radiologists with higher-quality draft reports compared to SEI \cite{sei}; 2) Our MLRG exhibits a certain ability to describe disease progression, as evidenced in case 2 with the statement ``the patient has received a right internal jugular vein catheter".

\textbf{Evaluation using large language models (LLMs).} Figure \ref{fig:5} illustrates the performance of our generated reports evaluated with GREEN \cite{2024-green}, a fine-tuned LLaMA 2 \cite{llama} designed to identify clinically significant errors and count matched findings. Results demonstrate that our MLRG outperforms R2Gen \cite{chen-etal-2020-generating}, CMN \cite{chen-etal-2021-cross-modal}, CGPT2 \cite{nicolson-improving-cvt2distilgpt2}, and SEI \cite{sei} in both \#Matched Findings and GREEN score, further confirming the advantage of our MLRG in generating coherent and clinically accurate radiology reports. For further details, please refer to the Appendix \ref{eva-large-language-models}.
\section{Conclusion}
In this paper, we introduced the MLRG method for chest X-ray report generation. We first proposed a multi-view longitudinal contrastive learning approach that leveraged the inherent spatiotemporal information from radiology reports to guide the pre-training of visual and textual representations. This approach not only captured differences among views but also flexibly extracted spatial features from current multi-view images and temporal features from longitudinal data, effectively leveraging spatiotemporal information in reports for pre-training. Subsequently, we presented a tokenized absence encoding technique to handle missing patient-specific prior knowledge. This technique allowed the multi-modal fusion network to adapt flexibly to scenarios with or without such data, ensuring the text generator can utilize available prior knowledge effectively. Extensive experiments on MIMIC-CXR, MIMIC-ABN, and Two-view datasets demonstrated that our MLRG outperforms existing SOTA methods in generating coherent and clinically accurate radiology reports, making it a strong contender for chest X-ray report generation. Future work will focus on using saliency maps \cite{2024-saliency-map} to learn region-based features and predict uncertainty \cite{uncertainty} to improve model reliability.

{
    \small
    \bibliographystyle{ieeenat_fullname}
    \bibliography{main}
}

\clearpage
\setcounter{page}{1}
\maketitlesupplementary

\appendix
\renewcommand{\thetable}{A\arabic{table}}
\renewcommand{\thefigure}{A\arabic{figure}}

\begin{table*}
\centering
\setlength{\tabcolsep}{1.4mm}
\begin{tabular}{ccccccccccccccc} 
\toprule
\multirow{3}{*}{\textbf{Observation}} & \multicolumn{7}{c}{\textbf{ MIMIC-CXR }} & \multicolumn{7}{c}{\textbf{ Two-view CXR }} \\ 
\cmidrule(r){2-8}\cmidrule(r){9-15}
 & \multirow{2}{*}{\textbf{ \%}} & \multicolumn{3}{c}{\textbf{SEI \cite{sei}}} & \multicolumn{3}{c}{\textbf{MLRG (Ours)}} & \multirow{2}{*}{\textbf{ \%}} & \multicolumn{3}{c}{\textbf{SEI \cite{sei}}} & \multicolumn{3}{c}{\textbf{MLRG (Ours)}} \\ 
\cmidrule(lr){3-5}\cmidrule(lr){6-8}\cmidrule(lr){10-12}\cmidrule(lr){13-15}
 &  & \textbf{P $\uparrow$} & \textbf{R $\uparrow$} & \textbf{F1 $\uparrow$} & \textbf{P $\uparrow$} & \textbf{R $\uparrow$} & \textbf{F1 $\uparrow$} &  & \textbf{P $\uparrow$} & \textbf{R $\uparrow$} & \textbf{F1 $\uparrow$} & \textbf{P $\uparrow$} & \textbf{R $\uparrow$} & \textbf{F1 $\uparrow$} \\ 
\midrule
ECM & 10.0 & \textbf{0.373} & 0.208 & 0.267 & 0.370 & \textbf{0.353} & \textbf{0.361} & 10.4 & 0.345 & 0.259 & 0.296 & \textbf{0.412} & \textbf{0.385} & \textbf{0.398} \\
Cardiomegaly & 14.8 & 0.599 & \textbf{0.633} & \textbf{0.616} & \textbf{0.629} & 0.570 & 0.598 & 14.4 & 0.578 & \textbf{0.602} & \textbf{0.589} & \textbf{0.627} & 0.550 & 0.586 \\
Lung Opacity & 13.8 & 0.519 & 0.170 & 0.256 & \textbf{0.594} & \textbf{0.317} & \textbf{0.413} & 13.6 & 0.526 & 0.197 & 0.287 & \textbf{0.549} & \textbf{0.295} & \textbf{0.384} \\
Lung Lesion & 2.5 & \textbf{0.462} & 0.021 & 0.041 & 0.429 & \textbf{0.046} & \textbf{0.082} & 3.0 & 0.179 & 0.030 & 0.051 & \textbf{0.297} & \textbf{0.045} & \textbf{0.078} \\
Edema & 8.3 & \textbf{0.526} & 0.361 & 0.428 & 0.516 & \textbf{0.448} & \textbf{0.480} & 6.5 & 0.420 & 0.368 & 0.392 & \textbf{0.457} & \textbf{0.455} & \textbf{0.456} \\
Consolidation & 3.3 & 0.218 & \textbf{0.194} & \textbf{0.205} & \textbf{0.259} & 0.150 & 0.190 & 3.1 & \textbf{0.296} & \textbf{0.192} & \textbf{0.233} & 0.204 & 0.115 & 0.147 \\
Pneumonia & 4.4 & 0.174 & 0.065 & 0.095 & \textbf{0.316} & \textbf{0.235} & \textbf{0.270} & 4.0 & 0.255 & 0.174 & 0.207 & \textbf{0.284} & \textbf{0.210} & \textbf{0.241} \\
Atelectasis & 10.9 & 0.469 & 0.395 & 0.429 & \textbf{0.499} & \textbf{0.475} & \textbf{0.487} & 10.0 & 0.457 & 0.425 & 0.440 & \textbf{0.496} & \textbf{0.444} & \textbf{0.469} \\
Pneumothorax & 1.0 & 0.174 & 0.039 & 0.064 & \textbf{0.426} & \textbf{0.230} & \textbf{0.299} & 0.7 & 0.417 & 0.109 & 0.172 & \textbf{0.457} & \textbf{0.291} & \textbf{0.356} \\
Pleural Effusion & 12.4 & 0.683 & \textbf{0.697} & \textbf{0.690} & \textbf{0.716} & 0.641 & 0.676 & 10.4 & 0.723 & \textbf{0.641} & \textbf{0.680} & \textbf{0.731} & 0.612 & 0.666 \\
Pleural Other & 1.6 & 0.167 & 0.022 & 0.039 & \textbf{0.231} & \textbf{0.054} & \textbf{0.087} & 1.9 & \textbf{0.250} & 0.071 & 0.111 & 0.194 & \textbf{0.083} & \textbf{0.116} \\
Fracture & 1.8 & 0.000 & 0.000 & 0.000 & \textbf{0.174} & \textbf{0.021} & \textbf{0.037} & 2.4 & 0.000 & 0.000 & 0.000 & \textbf{0.261} & \textbf{0.031} & \textbf{0.056} \\
Support Devices & 12.8 & 0.763 & 0.708 & 0.734 & \textbf{0.768} & \textbf{0.788} & \textbf{0.778} & 9.3 & \textbf{0.734} & 0.572 & 0.643 & 0.703 & \textbf{0.686} & \textbf{0.695} \\
No Finding & 2.4 & 0.161 & 0.597 & 0.253 & \textbf{0.233} & \textbf{0.629} & \textbf{0.340} & 10.3 & \textbf{0.509} & 0.899 & 0.650 & 0.490 & \textbf{0.933} & \textbf{0.643} \\ 
\cmidrule(lr){1-15}
micro avg & - & 0.523 & 0.410 & 0.460 & \textbf{0.549} & \textbf{0.468} & \textbf{0.505} & - & 0.522 & 0.447 & 0.481 & \textbf{0.532} & \textbf{0.474} & \textbf{0.501} \\
macro avg & - & 0.378 & 0.294 & 0.294 & \textbf{0.440} & \textbf{0.354} & \textbf{0.364} & - & 0.406 & 0.324 & 0.339 & \textbf{0.440} & \textbf{0.367} & \textbf{0.378} \\
\bottomrule
\end{tabular}
\caption{Clinical accuracy on the MIMIC-CXR and Two-view CXR datasets. ``ECM" refers to Enlarged Cardiomediastinum. ``P'', ``R'', and ``F1'' represent Precision, Recall, and F1 score, respectively.}
\label{atable:1}
\end{table*}

\begin{table*}
\centering
\begin{tabular}{cccccccc} 
\toprule
\multirow{2}{*}{\textbf{Observation}} & \multirow{2}{*}{\%} & \multicolumn{3}{c}{\textbf{SEI \cite{sei}}} & \multicolumn{3}{c}{\textbf{MLRG (Ours)}} \\ 
\cmidrule(r){3-5}\cmidrule(r){6-8}
 &  & \textbf{P $\uparrow$} & \textbf{R $\uparrow$} & \textbf{F1 $\uparrow$} & \textbf{P $\uparrow$} & \textbf{R $\uparrow$} & \textbf{F1 $\uparrow$} \\ 
\midrule
Enlarged Cardiomediastinum & 5.7 & 0.146 & 0.074 & 0.099 & \textbf{0.242} & \textbf{0.264} & \textbf{0.252} \\
Cardiomegaly & 12.7 & 0.515 & 0.627 & 0.566 & \textbf{0.547} & \textbf{0.785} & \textbf{0.644} \\
Lung Opacity & 20.2 & 0.640 & 0.342 & 0.446 & \textbf{0.649} & \textbf{0.512} & \textbf{0.572} \\
Lung Lesion & 5.0 & 0.333 & 0.035 & 0.063 & \textbf{0.357} & \textbf{0.052} & \textbf{0.090} \\
Edema & 7.1 & \textbf{0.464} & 0.524 & \textbf{0.492} & 0.441 & \textbf{0.547} & 0.489 \\
Consolidation & 3.3 & \textbf{0.359} & \textbf{0.383} & \textbf{0.371} & 0.354 & 0.270 & 0.306 \\
Pneumonia & 5.9 & 0.300 & 0.222 & 0.255 & \textbf{0.318} & \textbf{0.307} & \textbf{0.312} \\
Atelectasis & 10.5 & 0.381 & 0.441 & 0.409 & \textbf{0.445} & \textbf{0.578} & \textbf{0.503} \\
Pneumothorax & 0.0 & - & - & - & - & - & - \\
Pleural Effusion & 8.9 & 0.590 & 0.685 & 0.634 & \textbf{0.698} & \textbf{0.723} & \textbf{0.710} \\
Pleural Other & 3.2 & \textbf{0.158} & 0.056 & 0.082 & 0.135 & \textbf{0.081} & \textbf{0.101} \\
Fracture & 3.9 & 0.000 & 0.000 & 0.000 & 0.000 & 0.000 & 0.000 \\
Support Devices & 10.6 & 0.705 & 0.591 & 0.643 & \textbf{0.715} & \textbf{0.840} & \textbf{0.772} \\
No Finding & 3.0 & 0.175 & \textbf{0.540} & 0.265 & \textbf{0.262} & 0.466 & \textbf{0.335} \\
\cmidrule(lr){1-8}
micro avg & - & 0.466 & 0.408 & 0.435 & \textbf{0.513} & \textbf{0.517} & \textbf{0.515} \\
macro avg & - & 0.341 & 0.323 & 0.309 & \textbf{0.369} & \textbf{0.387} & \textbf{0.363} \\
\bottomrule
\end{tabular}
\caption{Clinical accuracy on the MIMIC-ABN dataset. ``P'', ``R'', and ``F1'' represent Precision, Recall, and F1 score, respectively.}
\label{atable:2}
\end{table*}

\begin{table*}
\centering
\begin{tabular}{ccccccccccc} 
\toprule
\multirow{2}{*}{\textbf{Generated Section(s)}} & \multicolumn{6}{c}{\textbf{NLG Metrics} $\uparrow$} & \multicolumn{4}{c}{\textbf{CE Metrics} $\uparrow$} \\ 
\cmidrule(r){2-7}\cmidrule(lr){8-11}
 & \textbf{B-1} & \textbf{B-2} & \textbf{B-3} & \textbf{B-4} & \textbf{MTR} & \textbf{R-L} & \textbf{RG} & \textbf{P} & \textbf{R} & \textbf{F1} \\ 
\midrule
FINDINGS & 0.411 & 0.277 & 0.204 & 0.158 & 0.176 & 0.320 & 0.291 & 0.549 & 0.468 & 0.505 \\
FINDINGS+IMPRESSION & 0.402 & 0270 & 0.197 & 0.152 & 0.172 & 0.327 & 0.289 & 0.558 & 0.468 & 0.509 \\
\bottomrule
\end{tabular}
\caption{Performance of generating ``FINDINGS" and ``IMPRESSION" sections on the MIMIC-CXR dataset.}
\label{atable:3}
\end{table*}

\begin{table*}
\centering
\setlength{\tabcolsep}{1.6mm}
\begin{tabular}{cccccccccc} 
\toprule
\multirow{2}{*}{\textbf{Method}} & \multicolumn{6}{c}{\textbf{\#Clinically Significant Errors $\downarrow$}} & \multirow{2}{*}{$\sum\nolimits_{j=(a)}^{(f)}{\rm{{\#Error}}}_{j}$ $\downarrow$} & \multirow{2}{*}{\textbf{\#Matched Findings $\uparrow$}} & \multirow{2}{*}{\textbf{GREEN $\uparrow$}} \\
\cmidrule(r){2-7}
& \textbf{(a)} & \textbf{(b)} & \textbf{(c)} & \textbf{(d)} & \textbf{(e)} & \textbf{(f)} & & &  \\ 
\midrule
R2Gen \cite{chen-etal-2020-generating} & 1.310 & 3.089 & \textbf{0.103} & \textbf{0.201} & \underline{0.082} & 0.142 & 4.926 & 1.803 & 0.283 \\
CMN \cite{chen-etal-2021-cross-modal} & 1.383 & 2.963 & \underline{0.127} & \underline{0.228} & \textbf{0.081} & 0.161 & 4.942 & 1.935 & 0.297 \\
CGPT2 \cite{nicolson-improving-cvt2distilgpt2} & \textbf{1.150} & 2.881 & 0.146 & 0.234 & 0.103 & 0.153 & \underline{4.666} & 1.967 & 0.313 \\
SEI \cite{sei} & 1.391 & \underline{2.610} & 0.154 & 0.273 & 0.108 & \underline{0.132} & 4.668 & \underline{2.101} & \underline{0.326} \\
MLRG (Ours) & \underline{1.277} & \textbf{2.469} & 0.199 & 0.284 & 0.091 & \textbf{0.130} & \textbf{4.451} & \textbf{2.261} & \textbf{0.353} \\
\bottomrule
\end{tabular}
\caption{Performance comparison of our MLRG and four baselines on the MIMIC-CXR test set in terms of ``\#Clinically Significant Errors" and ``\#Matched Findings". The best and second-best values are marked in \textbf{bold} and \underline{underlined}, respectively.}
\label{atable:4}
\end{table*}

\begin{figure*}
    \centering
    \includegraphics[width=1\linewidth]{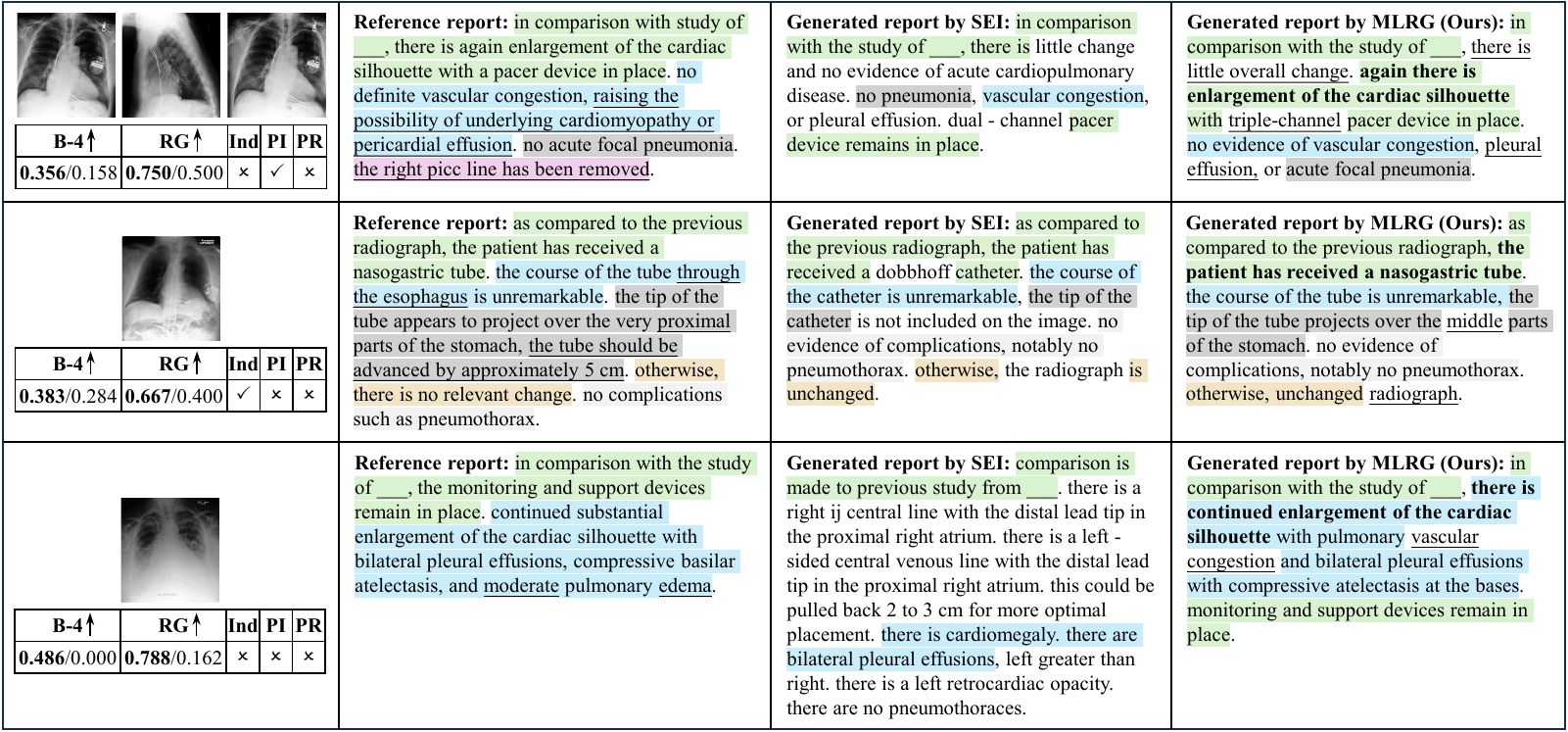}
    \caption{Examples of generated the ``FINDINGS" section on the MIMIC-CXR test set. Each ``A/B" cell corresponds to ``MLRG/SEI". Sentences from the reference report are highlighted in unique colors to clarify alignment with descriptions in the generated reports. Matching content in generated reports is shown in the same color. Correct temporal descriptions and failure descriptions of our MLRG are in \textbf{bold} and \underline{underlined}. ``Ind", ``PI", and ``PR" represent patient-specific indications, previous images, and previous reports, respectively.}
    \label{afig:1}
\end{figure*}

\begin{figure*}
    \centering
    \includegraphics[width=1\linewidth]{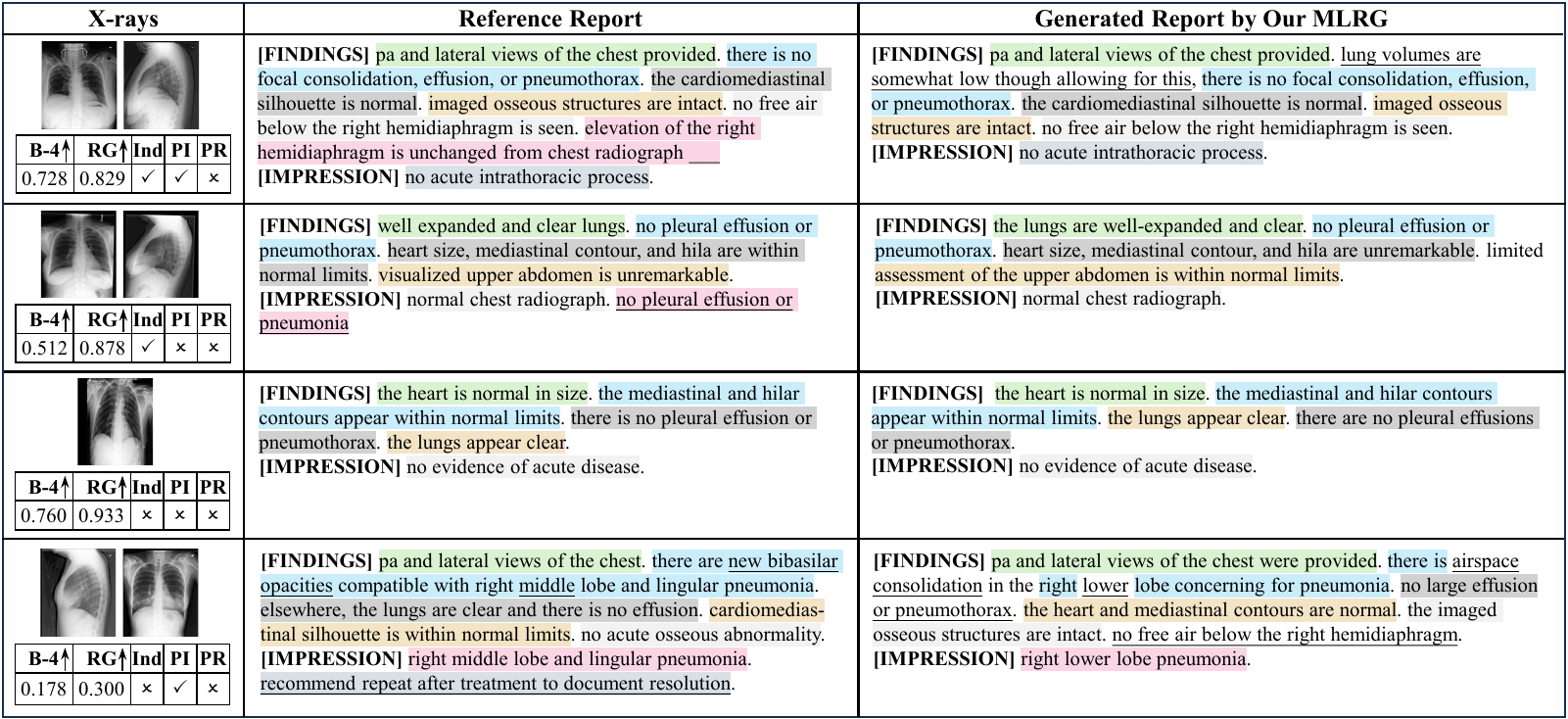}
    \caption{Generated examples of ``FINDINGS" and ``IMPRESSION" sections on the MIMIC-CXR test set. Sentences from the reference report are highlighted in unique colors to clarify alignment with descriptions in the generated reports. Matching content in generated reports is shown in the same color. Failure descriptions of our MLRG are \underline{underlined}. ``Ind", ``PI", and ``PR" represent patient-specific indications, previous images, and previous reports, respectively.}
    \label{afig:2}
\end{figure*}

\begin{figure*}
    \centering
    \includegraphics[width=1\linewidth]{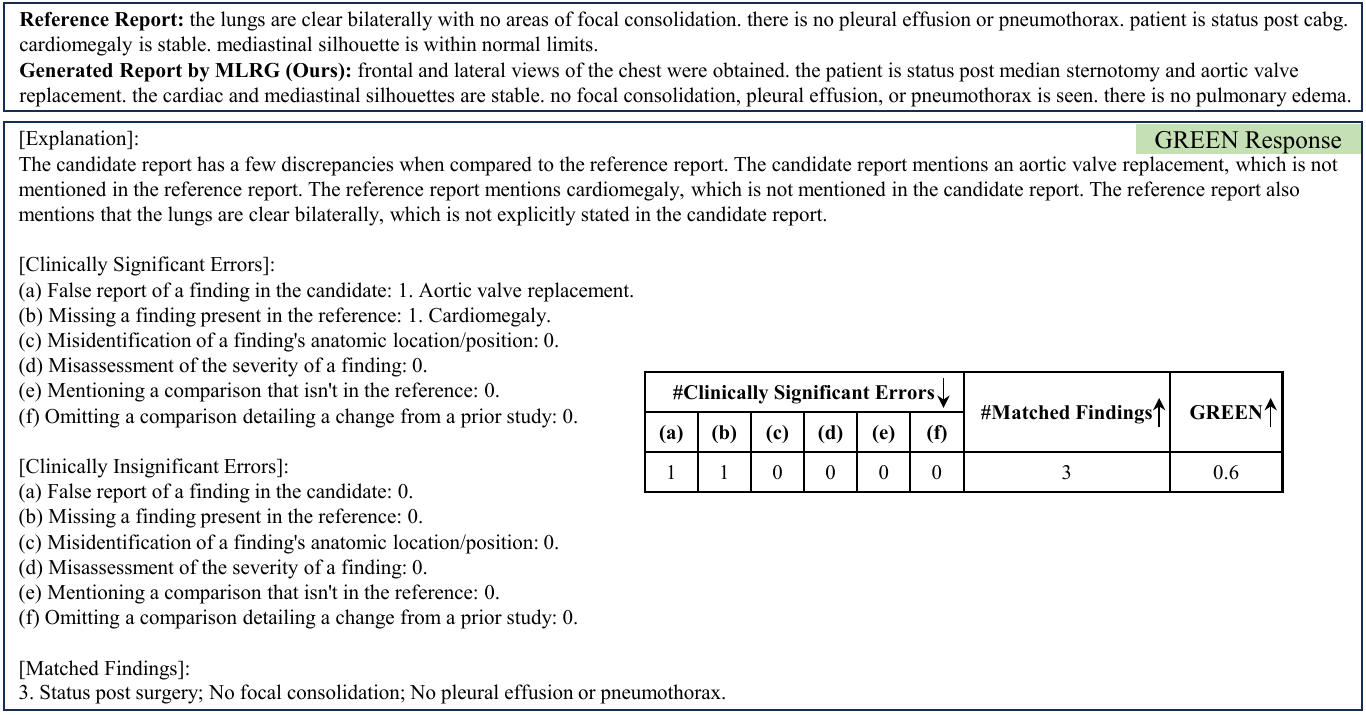}
    \caption{An output result of the GREEN model \cite{2024-green} on the MIMIC-CXR test set. ``\#Clinically Significant Errors" and ``\#Matched Findings" represent the number of clinically significant errors and matched findings, respectively.}
    \label{afig:3}
\end{figure*}


\section{Experiments}
\subsection{Implementation Details}
\label{implementation-details}
\textbf{1) MIMIC-CXR \cite{johnson-mimic-cxr-jpg}}: In stage 1, the model is trained for 50 epochs with a learning rate of 5e-5 and a batch size of 32. In Stage 2, we train MLRG for another 50 epochs, using a batch size of 14. The learning rate is set to 5e-6 for parameters from Stage 1 and 5e-5 for the remaining parameters. \textbf{2) MIMIC-ABN \cite{mimic-abn-ori} and Two-view CXR \cite{mcl}}: Since most images are derived from the MIMIC-CXR dataset, we directly fine-tune the model from Stage 2 on MIMIC-CXR, using a learning rate of 5e-6 and a batch size of 12. \textbf{3) Common settings}: Early stopping with a patience of 15 is employed to prevent overfitting. The ReduceLROnPlateau scheduler and the AdamW optimizer are applied. The natural language generation (NLG) metrics are calculated with the pycocoevalcap\footnote{https://github.com/tylin/coco-caption}. For clinical efficacy (CE) metrics, Precision, Recall, and F1 score metrics are computed using the f1chexbert\footnote{https://pypi.org/project/f1chexbert/} library, and the F1 RadGraph metric is calculated with the radgraph\footnote{https://pypi.org/project/radgraph/} library.

\subsection{Clinical Accuracy of 14 Observations}
Tables \ref{atable:1} and \ref{atable:2} show the clinical accuracy of 14 observations annotated by CheXpert \cite{irvin-chexpert}  on the MIMIC-CXR, MIMIC-ABN, and Two-view CXR datasets. Results show that our MLRG outperforms SEI \cite{sei} on most observations. Even though MLRG is not specifically tailored for imbalanced observations, it still slightly surpasses the baseline on challenging observations like \textit{Pneumothorax} and \textit{Pleural Other}.

\subsection{Performance of Generating ``FINDINGS'' and ``IMPRESSION'' Sections}
Radiology reports typically consist of three key sections: ``INDICATION", which outlines the visit reasons or symptoms; ``FINDINGS", which details observations from current multi-view images and comparisons with the patient's medical history; and ``IMPRESSION", which summarizes the key conclusions or diagnostic interpretations based on the ``FINDINGS". Table \ref{atable:3} presents the performance of generating ``FINDINGS" and ``IMPRESSION" sections, with specific examples in Figure \ref{afig:2}. Since most existing methods focus primarily on generating the ``FINDINGS" section, peer methods are not included in Table \ref{atable:3}. The results indicate that our MLRG is capable of generating both sections with minor modifications. Specifically, we utilize special tokens, ``[FINDINGS]" and ``[IMPRESSION]", before the respective section content to distinguish between them. These sections are then combined to form the final reference reports, with all other settings remaining identical to those used for the ``FINDINGS" generation.

\subsection{Qualitative Analysis}
Figure \ref{afig:1} provides additional examples of the ``FINDINGS" section generated by SEI \cite{sei} and our MLRG, while Figure \ref{afig:2} presents examples of both the ``FINDINGS" and ``IMPRESSION" sections from MLRG. These results suggest that 1) Our MLRG is highly competitive in generating both ``FINDINGS" and ``IMPRESSION" sections, as well as the ``FINDINGS" section alone, for chest X-ray reports. 2) MLRG still has room for improvement in describing lesion attributes. For example, in Figure \ref{afig:1}, MLRG incorrectly describes the ``proximal parts of the stomach" as the ``middle parts". This occurs because MLRG has not fully learned region-level features. To improve this, we are exploring the use of saliency maps \cite{2024-saliency-map} to enhance regional feature learning and the accuracy of lesion descriptions.

\subsection{Evaluation Using Large Language Models}
\label{eva-large-language-models}
Inspired by \cite{yu2023evaluating}, the GREEN model \cite{2024-green} identifies six categories of clinical errors: (a) False report of a finding in the candidate; (b) Missing a finding present in the reference; (c) Misidentification of a finding's anatomic location/position; (d) Misassessment of the severity of a finding; (e) Mentioning a comparison that isn't in the reference; (f) Omitting a comparison detailing a change from a prior study. The GREEN score for the $i^{th}$ sample is defined as:
\begin{align}
{\rm{GREEN}}_i = \frac{{{\rm{\#Matched}}\;{\rm{Findings}}_i}}{{{\rm{\#Matched}}\;{\rm{Findings}}_i + \sum\nolimits_{j=(a)}^{(f)} {\rm{{\#Error}}}_{i,j} }},
\end{align}
where $\sum\nolimits_{j=(a)}^{(f)} {\rm{{\#Error}}}_{i,j}$ represents the total clinically significant errors for the $i^{th}$ sample across categories (a) to (f). ``\#Matched Findings" denotes the number of matched findings between generated and reference reports. Figure \ref{afig:3} illustrates the GREEN model's output on the MIMIC-CXR test set. Furthermore, we compare our MLRG with R2Gen \cite{chen-etal-2020-generating}, CMN \cite{chen-etal-2021-cross-modal}, CGPT2 \cite{nicolson-improving-cvt2distilgpt2}, and SEI \cite{sei} in terms of ``\#Clinically Significant Errors" and ``\#Matched Findings", as summarized in Table \ref{atable:4}. The results reveal the following: 1) Our MLRG achieves the highest ``\#Matched Findings" and GREEN score, with the fewest total clinically significant errors. This further confirms the effectiveness of our MLRG in generating clinically accurate radiology reports. 2) MLRG performs best in category “(f) Omitting a comparison detailing a change from a prior study”, suggesting its ability to effectively extract temporal features from multi-view longitudinal data. 3) Although MLRG shows higher error counts than the baselines in categories (c) and (d), its total clinically significant errors across categories (a) to (f) remain lower than those of all baselines. To improve the accuracy of severity assessments and anatomical location descriptions, we are exploring the integration of saliency maps \cite{2024-saliency-map} and MIMIC-CXR-VQA \cite{mimic-cxr-vqa-nips} data to learn region-based features, aiming to generate more precise descriptions of findings.

\end{document}